\documentclass{article}

\usepackage{microtype}
\usepackage{graphicx}
\usepackage{subfigure}
\usepackage{booktabs} %

\usepackage{url}

\usepackage{breakurl}
\usepackage[breaklinks]{hyperref}

\usepackage[accepted]{icml2024}

\usepackage{amsmath}
\usepackage{amssymb}
\usepackage{mathtools}
\usepackage{amsthm}

\usepackage{tikz}
\usepackage{pgfplots}

\usepackage[capitalize,noabbrev]{cleveref}

\theoremstyle{plain}

\theoremstyle{definition}

\theoremstyle{remark}

\usepackage[textsize=tiny]{todonotes}

\icmltitlerunning{The Dilemma of Uncertainty Estimation for General Purpose AI in the EU AI Act}

\begin{document}

\twocolumn[
\icmltitle{The Dilemma of Uncertainty Estimation for General Purpose AI in the European Union Artificial Intelligence Act}

\icmlsetsymbol{equal}{*}

\begin{icmlauthorlist}
\icmlauthor{Matias Valdenegro-Toro}{equal,rugfse}
\icmlauthor{Radina Stoykova}{equal,ruglaw}
\end{icmlauthorlist}

\icmlaffiliation{rugfse}{Department of AI, University of Groningen}
\icmlaffiliation{ruglaw}{Transboundary Legal Studies, University of Groningen}

\icmlcorrespondingauthor{Matias Valdenegro-Toro}{m.a.valdenegro.toro@rug.nl}
\icmlcorrespondingauthor{Radina Stoykova}{r.stoykova@rug.nl}

\icmlkeywords{Machine Learning, ICML, Artificial Intelligence Act (AI act), Accuracy, Uncertainty Estimation, General Purpose AI Models, Systemic Risk}

\vskip 0.3in
]

\printAffiliationsAndNotice{\icmlEqualContribution} %

\begin{abstract}
The AI act is the European Union-wide regulation of AI systems. It includes specific provisions for general-purpose AI models which however need to be further interpreted in terms of technical standards and state-of-art studies to ensure practical compliance solutions. This paper examines the AI act requirements for providers and deployers of general-purpose AI and further proposes uncertainty estimation as a suitable measure for legal compliance and quality assurance in training of such models.
We argue that uncertainty estimation should be a required component for deploying models in the real world, and under the EU AI Act, it could fulfill several requirements for transparency, accuracy, and trustworthiness. However, generally using uncertainty estimation methods increases the amount of computation, producing a dilemma, as computation might go over the threshold ($10^{25}$ FLOPS) to classify the model as a systemic risk system which bears more regulatory burden.
\end{abstract}

\section{Introduction}
The AI act (AIA) is the first comprehensive regulation of AI systems in the European Union, that was formally signed in June 2024 \cite{AIact2024}. It is expected to enter into force in 2025. Specific attention in the negotiations for AIA was given to transparency and model evaluation obligations for providers and deployers of general-purpose AI models (GPAI). 
The legislator considered that GPAI can have significant risks to society and fundamental rights. When such models under perform this can lead to negative consequences for individuals which vary from enforcing stereotypes in society to triggering legal consequences and public safety concerns. For example, Chat GPT was often discussed in terms of gender and racial bias, \cite{exampleBias} as well as its inability to filter potentially dangerous prompts for mixing poisons or explosives. A notorious case in US involved an attorney who used Chat GPT to prepare a filing for a civil case and ended up citing non-existing case law due to hallucinations in the model. \cite{exampleLAwyer} In another example, a Canadian airline company was forced to honor a refund policy which was hallucinated by the company`s chat bot. \cite{exampleAIR} 

Therefore, the AI act specifies concrete transparency documentation and model evaluation requirements for GPAI with particular focus on metrics to evaluate the model such as accuracy and performance metrics, quality of datasets assurance, and robustness against errors. Apart from these general accountability requirements, the AIA relies on multi-stakeholder cooperation between industry, academia, and standardisation bodies to establish concrete standards, technical specifications, and best practices for testing and evaluation of AI systems which will support the implementation of the AIA.

In this paper, we propose uncertainty estimation as a standard measure for GPAI. We argue that to enable general-purpose AI (GPAI) models evaluation and human oversight and to ensure legal compliance, it would be useful for providers and deployers of AI systems to be informed on the model confidence in output. For a human it is natural that she can express full confidence, partial confidence, or reply with a simple "I don't know" \cite{barrett2023ai}. Similarly, it should be expected that AI models can perform the task of confidence estimation themselves as well, as this information is useful for developers and deployers to give a weight to AI model responses, and as a proxy measure about really trusting the prediction, same as with other human-generated opinions and documents.

As development of trustworthy AI models is a core principle in AI act, this paper proposes and studies the feasibility of uncertainty estimation as a mandatory component of training and evaluation of AI models, as it is currently not widely adopted and AI model developers do not generally build models with these advanced capabilities.

However, GPAI providers might be reluctant to implement such measure as its use during training will increase the amount of computation for the model development which will require compliance with the more stringent regime of GPAI with systemic risk. Therefore, we further examine the benefits, computational costs, and limitations of the measure as a GPAI quality benchmark.

This paper is structured as follows: Section 1 introduced the topic, while Section 2 focuses on background and methods for uncertainty estimation in AI. Section 3 summarises the most important legal provisions for regulation of GPAI. In Section 4 we discuss the feasibility of uncertainty estimation as a benchmark for quality assurance and legal compliance with the AIA regulation, while Section 5 provides discussion and conclusions.   

\section{Uncertainty in Machine Learning}
The power of GPAI models is that they can make predictions of many kinds, but these predictions are only beneficial if they are approximately correct \cite{campos2023definition}. Incorrect predictions, or popularly known as "hallucinations" are defined as "generated content that is nonsensical or unfaithful to the provided source content" \cite{ji2023survey} indicating that they are simply wrong outputs. \cite{huang2023survey} Hallucinations result from data or modelling problems and are not useful predictions to a human given some context or prompt. Determining truth in AI models is very difficult as these models are not trained to produce an objective "truth", but to reproduce tokens from the training set, which make more or less meaningful answers, but there are no guarantees for correctness. 
The overall concept of estimating AI confidence is the field of uncertainty estimation in machine learning, and there are many techniques for this purpose, relying on different assumptions \cite{gawlikowski2023survey}. The overall issue with this field is that estimating AI model confidence usually requires additional computational resources, and it needs to be explicitly considered during the training process.

Large AI models like Large Language Models and Vision-Language Models often do not have proper confidence estimation capabilities \cite{groot2024overconfidence}, by outputting confidences that are not a reflection of true confidence, as correct and incorrect answers have similar high confidences, and this prevents discrimination of correct and incorrect predictions \cite{huang2023look} \cite{xiong2023can}. The overall concept of confidence estimation requires that incorrect predictions have lower confidence than correct predictions, ideally with incorrect predictions having 0\% confidence, and correct predictions having 100\% confidence.

\subsection{Methods for Uncertainty Estimation}
Methods to estimate uncertainty for machine learning models can be broadly divided into two categories: direct methods like ensembles that directly provide uncertainty estimates, and sampling methods like MC-Dropout, where forward passes of the model correspond to samples of a posterior probability distribution. In both kinds of methods, samples or forward passes are combined to build an output probability distribution.

\begin{equation}
    \mu(x) = M^{-1} \sum^M_i \text{model}_i(x)
    \label{eq:pred_mean}
\end{equation}
\begin{equation}
    \sigma^2(x) = M^{-1} \sum^M_i [\text{model}_i(x) - \mu(x)]^2
    \label{eq:pred_var}
\end{equation}

Where $M$ is the number of forward passes or models in the ensemble, and $\text{model}_i$ represents the predictions of the $i$-th model in the ensemble or the $i$-th forward pass sample.

The variance of the predicted probability distribution $\sigma^2(x)$ is a measure of uncertainty, the larger the variance, the more uncertain the prediction is, and more likely to be incorrect. The mean of the predicted probability distribution $\mu(x)$ corresponds to the combined prediction that is given to the end user.

\textbf{Direct Methods}. The most popular method is Ensembles, where any neural network is trained $M$ times on the same dataset, and due to random weight initialization, the model converges to different weights. At inference time, each model in the ensemble (the $\text{model}_i$) makes a prediction and they are combined using Eq \ref{eq:pred_mean} and \ref{eq:pred_var}. A typical value is $M = 5$.

\textbf{Sampling Methods}. Monte Carlo Dropout is a popular sampling technique, where Dropout layers are inserted in the neural network architecture, but these layers are active both during training and inference, and the random neuron dropping effect of Dropout is enabled when making predictions, producing stochastic outputs. The output distribution is reconstructed using Eq \ref{eq:pred_mean} and \ref{eq:pred_var} via $M$ forward passes of the network, with a typical value $M \in [10, 50]$.

\textbf{Other Methods}. There are methods that use a single network architecture, avoiding the need for ensembles of multiple networks or costly sampling. Deterministic Uncertainty Quantification (DUQ) uses a radial basis function output layer to encode per-class centroid \cite{van2020uncertainty}, while Deep Deterministic Uncertainty uses an lipschitz regularized ResNet that preserves distances to enable feature space density estimation \cite{mukhoti2023deep}. The disadvantages of these methods is that they are not general and make assumptions, for example only being defined for classification tasks, and still they require a slight increase ($\sim 10\%$) in computation.

\subsection{Computational Requirements}

In the previous section, we argued that using uncertainty estimation methods requires changes to the training process of the model, but more fundamentally, it also changes the prediction process. Additional computation in the form of ensemble models or multiple forward passes are often required, increasing the computational costs of applying uncertainty estimation methods to machine learning models, in comparison with not applying these techniques.

To make predictions with uncertainty, multiple forward passes or multiple models are required, which increases their computational cost linearly as a function of $M$, compared over the original single model.

\begin{figure}
    \centering
    \begin{tikzpicture}
	   \begin{axis}[domain=1:10, ylabel=Uncertainty Quality, xlabel=Compute (FLOPs), ticks=none]
           \addplot+[mark=none] {3.5 / (1 + exp(-0.6*x))};
           \addplot+[mark=none] {3.0 / (1 + exp(-0.7*x))};
           \addplot+[mark=none] {2.5 / (1 + exp(-0.9*x))};
	   \end{axis}
    \end{tikzpicture}
    \caption{Sample relationship between computation and uncertainty quality. Note how most methods plateau, and these are synthetic results representative of real-world}
    \label{fig:synth_compute_uncertainty}
\end{figure}
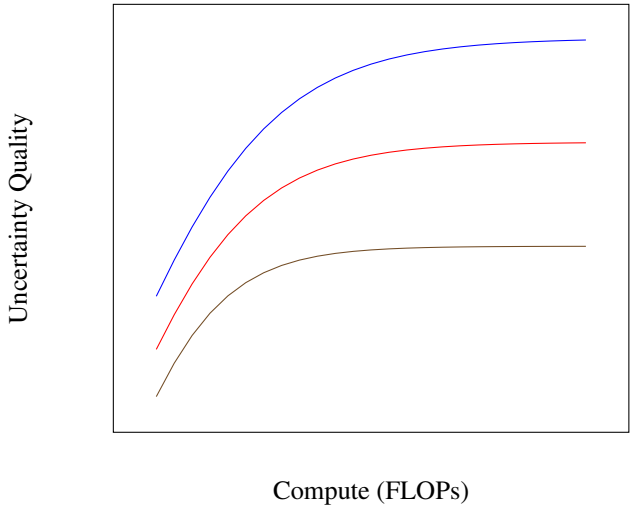

A typical value for ensemble models is $M = 5$. The selection of $M$ provides a trade-off between uncertainty quality and computational requirements. More computation allows for more forward passes or models (larger $M$) and better uncertainty quality, but this can become computationally expensive to compute. Figure \ref{fig:synth_compute_uncertainty} shows this concept, where uncertainty quality indicates the ability to separate correct from incorrect predictions in various settings.

Another disadvantage of uncertainty estimation methods is, since they change the training process, sometimes depending on the method, performance on the task itself (classification or regression) can change, either decreasing or increasing.

\section{GPAI in the context of the European Artificial Intelligence Act}
According to Art 3 (63) AIA, an AI model is defined as general purpose if: \textit{(1)} it is trained with a large amount of data \textit{(2)} uses self-supervision at scale \textit{(3)} displays significant generality and \textit{(4)} is capable of competently performing a wide range of distinct tasks. Currently this definition encompasses two types of AI models: generative AI and foundation models. Generative AI refers to deep learning that generates content like text, video, images or code depending on the provided input. Foundation models on the other hand are general purpose or widely applicable models for many tasks, which does not necessarily imply generating data.
 
In addition, if GPAI is used in specific sectors or for the tasks listed in Art. 6 and Annex III e.g., education, law enforcement, employment, the GPAI models may be classified as high-risk AI by themselves or as component of other high-risk AI system. The use of GPAI in high-risk systems is a separate compliance issue that needs to be discussed in detail. Nevertheless, we examine the stringent regime for GPAI classified or part of high-risk AI system only in the context of uncertainty estimation and its feasibility as legal compliance measure. This is desirable also because the AIA recommends voluntary application of some or all of the mandatory requirements applicable to high-risk AI systems.   

Further the AI act classifies two groups of GPAI models based on compute threshold: GPAI and GPAI with systemic risks. The GPAI is considered with systemic risk or high impact capabilities if the cumulative amount of compute used for model`s training is greater than $10^{25}$ floating point operations (FLOPs) (Art. 51 (2) AIA). In this paper we focus on interpreting the new regulatory requirements for GPAI and specifically on large language models like ChatGPT v.4 or multi-modal models (audio, video, text, etc) which fit the definition of GPAI with systemic risk.
 
The AI act specifies concrete transparency documentation and model evaluation requirements for providers and deployers of general-purpose AI models (GPAI) in Art. 53 with specific focus on metrics to evaluate the GPAI model such as accuracy and performance evaluation metrics, quality of datasets assurance, and robustness against errors (explicitly listed in Annex XI AIA). Moreover, the AIA requires human oversight measures, which enable humans to interpret the AI system output and if needed to intervene in order to avoid negative consequences or risks, or stop the system if it does not perform as intended. This can significantly improve their integration in AI systems for specific tasks as it requires also close cooperation with downstream providers (those who implement the GPAI model in their own AI systems). 

GPAI with systemic risk is a model that has high-impact capabilities in the sense that it is characterised with unpredictability, emerging capabilities, and continuous learning. Therefore, the legislator considers that GPAI with systemic risk can have actual or reasonably foreseeable negative effects on public health, safety, public security, fundamental rights, or the society as a whole, that can be propagated at scale (Art. 3(65) AIA).

Providers of such models must follow a stringent regime with additional obligations to perform mandatory model evaluation and adversarial testing according to specific standards, which ensure assessment and mitigation of systemic risks. (Art. 55-56 AIA)  The newly established AI Office is a body that will facilitate the development of standards and codes of practice. Therefore, research and discussion on what practical measures are necessary to assess the risks of GPAI is of crucial importance in this regulatory initiative. Despite the strict legal regime, existing guidelines on risk management in GPAI report issues with high uncertainty and lack of standards to mitigate it \cite{barrett2023ai}, while an initial assessment by Stanford on the most common Gen AI models show that they suffer from data quality and governance issues, lack of transparency, and low robustness. \cite{bommasani2023eu-ai-act} 

\section{Feasibility of Uncertainty Estimation as a Measure for AIA Compliance}
The AIA is a framework law as it provides for a general accountability regime for AI systems, but relies on industry, researchers and other stakeholders to develop further best practices and standards for operationalization of the AIA in the concrete domain and type of AI systems. Further, we examine firstly how uncertainty estimation can benefit compliance with the high-risk AIA requirements, and secondly for its feasibility for GPAI and GPAI with systemic risk quality assurance.

\subsection{Uncertainty estimation to Support High-risk GPAI Assessment}
GPAI models that are implemented in AI systems for domains and tasks that present high risk to safety and fundamental rights of individuals (see Art. 6 and Annex III) are obliged to mandatory comply with the high-risk requirements in Chapter II of the act summarized in
Table \ref{tab:aiact}

\begin{table}[!h]
\centering
\small
\begin{tabular}{p{3cm}p{4.4cm}}
    \toprule
        \textbf{AIA High-risk requirements} & \textbf{Description} \\
        \midrule
		 Risk-management system (Art. 9) & Identification, testing, and mitigation of foreseeable risks for the intended purpose\\ 
		 Data and data governance (Art. 10) & High-quality training, validation, and testing data (relevant; representative, accurate) with application-area specific properties; identifying individuals or groups affected \\
		Technical documentation (Art. 11) & Extensive; includes e.g. general description of the AI system; the elements of the AI system and of the process for its development; the validation and testing procedures used; about the monitoring, functioning and control of the AI system; the risk management system. \\
		 Record-keeping (Art. 12) & Documentation and logs to ensure accountability and transparency \\
		 Transparency information (Art. 13) & Appropriate degree of transparency to enable users to interpret the system’s output and use it \\
		 Human oversight (Art. 14) & Built-in or user-implemented measures to allow a person to correctly interpret the AI high-risk system’s output \\
        Accuracy, robustness, and cybersecurity (Art. 15) & Resilience as regards errors, faults or inconsistencies, including protection against discrimination and adversarial machine learning \\
        \bottomrule
\end{tabular}
\caption{AIA Requirements for High-Risk AI Systems \label{tab:aiact}} 
\end{table} 

\subsubsection{Supporting Risk Management}

The first requirement for high risk GPAI is to be accompanied with risk management system that is maintained and updated throughout the GPAI entire life cycle. Such system should encompass two types of measures: \textit{(i)} for the identification, analysis, and mitigation of foreseeable risks (Art. 9 (2-5)); and  \textit{(ii)} for the testing of most appropriate risk management measures (Art. 9 (6-8)). 

Providers and deployers of GPAI should always consider as foreseeable risk that the GPAI model can make incorrect predictions. 

GPAI uncertainty estimation can support this objective as it provides a way to identify and record hallucinations and incorrect predictions by providing a threshold on model confidence to separate correct from incorrect predictions. Further, the predictions that are bellow this threshold can be analysed to identify the origin of the errors and possible mitigation strategies. As the measure shows the probability of the prediction being correct, it also allows to detect possible misuse of the system. For example, if currently ChatGPT can be tricked to generate fake news, with uncertainty estimation, the model will provide proof that the output might be incorrect. Interestingly, the legislator explicitly stated in recital 65 that addressing foreseeable misuse of AI system should not require specific additional training. To the contrary, to the best of our knowledge additional training is always required when misuse or risk mitigation strategies are employed.

\subsubsection{Improvement of Dataset Governance}
Current practices of adding just more data to train GPAI models were efficient in improving performance of the model to certain extend, but eventually if such data is of poor quality the model degrades over time. Therefore, the AIA considers that the performance of GPAI models depends on the quality of the datasets used for training, validation, and testing. Art. 10 AIA defines concrete data management practices and stringent requirements for quality and relevance of datasets. The origin of data, relevance, representativness, and data preparation techniques must be clearly stated, while mitigation of errors or biases in the data should be demonstrated. However, a preliminary EU study concluded that complience with those requirements might be challenging in practice, as currently there are no universally agreed standard for dataset quality assurance, while the quality of data is domain and AI system specific. \cite{de_miguel_beriain_auditing_2022}

Uncertainty measures for GPAI can assist in fulfilling partly data governance requirements. It improves the quality of the training process as it allows for the system to detect and report by itself inaccurate predictions (low confidenc). In this sense, the uncertainty estimation measure allows to minimise the negative effect of low-quality data on the systems output and potentially to trace the reasons for high uncertainty thresholds and to curate the datasets further.

Data uncertainty (also known as aleatoric uncertainty) in labels can be estimated by a model if trained with an appropriate setup, and then the model can report data (aleatoric) and model (epistemic) uncertainty separately \cite{valdenegro2022deeper}, which have different meanings. High model uncertainty reports gaps in the training set and inputs far from the training set distribution, while high data uncertainty reports problems with labels, such as ambiguous or incorrect labels.

\subsubsection{Enabling Documentation, Transparency, and Human Oversight}
Art. 11 and 12 of the AIA require sufficient technical documentation and record keeping for 
which are both measure to enable more transparency and human oversight in high-risk GPAI. Annex IV AIA specifies concrete requirements for technical documentation, where uncertainty estimation measure can be used to satisfy several of them as follows:
\begin{itemize}
    \item (1)(b) how AI system can be used to interact with 
    \item (1)(c) the computational resources used to develop, train, test and validate the AI system
    \item (1)(e) technical measures needed to facilitate the interpretation of the outputs of AI systems
    \item (1)(f) he technical solutions adopted to ensure continuous compliance of the AI system with AIA
\end{itemize}
Confidence level estimation for GPAI can assist providers and deployers to understand and assess the reliability of the GPAI output (art. 14 AIA). In particular, such measure will allow to find a low confidence answers that are likely to be incorrect or present inputs that were unexpected during training, which can then be logged and used to improve the system in a next iteration.

\subsubsection{A Mmeasure to Assess GPAI Accuracy and Robustness Against Errors}

AIA requires accuracy and robustness measurements as well as continuous performance evaluation of the AI system in order to ensure resilience against limitations of the AI system such as errors, faults or inconsistencies and sufficient transparency information for deployers of AI systems regarding such limitations (see Art. 15 (3) and Art. 13 (3)(b)(ii)). The desired level of accuracy depends on the domain and the level of error tolerance for the specific task. For example medical AI applications needs to have high accuracy across multiple population groups and proper confidence estimation for physicians to trust predictions and take a deeper look on low confidence predictions, while leisure applications not need to have high accuracy, as it is a low stakes setting.

A limitation for the currently used performance metrics for GPAI is that they report on overall model accuracy, but does not account for the nature, origin, or severity of reported error rates. One very clear example is in face recognition algorithms \cite{buolamwini2018gender}, where performance in terms of accuracy decreases significantly for darker skin tones, as they are less prevalent in the training set. Without proper validation, biases in a model can go unchecked.

Uncertainty estimation can be used as a per-sample proxy for accuracy metric for GPAI as well as a source to examine the nature and severity of errors. For example, in the case of face recognition, when the model makes incorrect predictions, these should have a high uncertainty or low confidence, and this should be examined by a human, by setting threshold on uncertainty.

An important case are hallucinations, which should also be predictions with low confidence, which can then be logged, and even a GPAI system can reject to produce an answer instead of showing a hallucination to the end user.

\begin{table*}[!ht]
    \begin{tabular}{p{0.08\linewidth}p{0.45\linewidth}p{0.40\linewidth}}
		\toprule
		EU AI Act Article & Uncertainty Use & Dilemma\\
		\midrule
		Art 9  & Model will make incorrect predictions, model confidence can help detect incorrect predictions and "hallucinations" & No guarantees for uncertainty quality under distribution shifts\\
		Art 10 & Detect low quality data & Further steps necessary, no guarantees\\
		Art 11 & Use of uncertainty estimation methods should be documented, calibration plots and errors reported. & AI developers can choose to avoid using uncertainty methods due to more computational needs\\		
		Art 12 & Prediction confidence should be logged for human evaluation & More information to be stored and interpreted, increased energy use\\
		Art 13 & Model confidence gives information about trustworthiness of a prediction & Additional computation and energy use for high quality uncertainty\\
		Art 14 & A human can use model confidence to detect incorrect predictions or system misuse & Additional computation and energy use for high quality uncertainty\\
		Art 15 & Errors and faults can be detected with uncertainty estimation, particularly out of distribution inputs. Models with UQ are often more robust & Additional computation and energy use for high quality uncertainty\\
		\bottomrule
	\end{tabular}
    \caption{Summary of EU AI Act vs Uncertainty Use Cases and their Dilemmas}
    \label{tab:dilemma_summary}
\end{table*}

\subsection{Limitations}

The use of uncertainty estimation methods come with many limitations. In general there are no guarantees on quality of uncertainty \cite{ovadia2019can}, meaning that incorrect predictions can still have high certainty, and there is much research to improve calibraton of machine learning models.

Part of our main argument is that GPAI systems with uncertainty estimation require more energy use, during both training and at prediction time, and this is a major limitation, as the public and regulators would like to reduce the energy consumption of GPAI systems.

Finally, uncertainty estimation methods do not directly address bias in datasets or GPAI systems, which are built indirectly by humans, and for this purpose other methods must be used, coming from the literature in fairness of machine learning algorithms.

\section{Discussion and Conclusions}
The new AIA act is a brave fist step towards a comprehensive accountability regime for AI systems in general, and GPAI in particular. However, the act is a framework law that requires its interpretation with respect to each AI model or system on case-by-case bases. AIA also relies on the development of common standards and technical specifications to establish best practices for compliance with the regulation. One standard proposed and examined in this paper for its feasibility is uncertainty estimation. Providers and deployers of GPAI should  know if the output they obtain from the system is correct or they should trust the prediction, but current GPAI systems do not give confidence estimates. This paper provided arguments that GPAI models should be trained with proper uncertainty estimation methods, and provide confidence estimates to the end user.

We demonstrated that uncertainty estimation measure is a practical solution for compliance with AIA requirements for transparency, technical documentation, robustness, and human oversight as it allows providers and developers to disregard erroneous output and further examine and curate the models data to mitigate hallucination problems. A summary of our proposed use cases are presented in Table \ref{tab:dilemma_summary}.

Some controversies emerge in the field of GPAI since integration of legal compliance measures like uncertainty estimation also increases the FLOPs for model training. The legislator approach to decide if GPAI poses systemic risk based on the amount of compute is a good starting point, but it is a simplistic view, as computation can be used for different purposes that might not imply emerging or unexpected properties of a model. This presents a dilemma under the AIA. It seems that the legislator considers more computations for GPAI training as an indicator for increased risk of the system, to the contrary, we demonstrated that measure to ensure evaluation of the model and legal compliance can increase the computations in order to reduce the risks.

It is questionable, if methods to increase legal compliance that also increase the computational and energy consumption for the AI system should be encouraged and if so should those computations be excluded from the FLOPs count in order to avoid the classification of the system as systemic risk. This presents a legal dilemma that might discourage developers from implementing advanced model performance methods like uncertainty estimation.

\clearpage
\bibliography{biblio}
\bibliographystyle{icml2024}

\end{document}